\documentclass{ifacconf}
\usepackage{graphicx}      
\usepackage{natbib}        
\usepackage{amsmath}
\usepackage{titlesec}
\usepackage{xcolor}

\usepackage{hyperref}
\begin{document}
\begin{frontmatter}

\title{Object Manipulation in Marine Environments using Reinforcement Learning } 
\thanks[footnoteinfo]{Sponsor and financial support anonymous.}

\author{Ahmed Nader}, 
\author{Muhayy Ud Din}, 
\author{Mughni Irfan}, and
\author{Irfan Hussain}
\address{Khalifa University Center for Autonomous Robotic Systems - Khalifa University, UAE.}



\begin{abstract}

Performing intervention tasks in the maritime domain is crucial for safety and operational efficiency. The unpredictable and dynamic marine environment makes the intervention tasks such as object manipulation extremely challenging. 
This study proposes a robust solution for object manipulation from a dock in the presence of disturbances caused by sea waves.
To tackle this challenging problem, we apply a deep reinforcement learning (DRL) based algorithm called Soft.
Actor-Critic (SAC). SAC employs an actor-critic framework; the actors learn a policy that minimizes an objective function while the critic evaluates the learned policy and provides feedback to guide the actor-learning process. We trained the agent using the PyBullet dynamic simulator and tested it in a realistic simulation environment called MBZIRC maritime simulator. This simulator allows the simulation of different wave conditions according to the World Meteorological Organization (WMO) sea state code. Simulation results demonstrate a high success rate in retrieving the objects from the dock. The trained agent achieved an 80 percent success rate when applied in the simulation environment in the presence of waves characterized by sea state 2, according to the WMO sea state code.

\end{abstract}

\end{frontmatter}

\section{Introdcution}



Marine robotics has gained significant importance in recent years, where autonomous systems are increasingly used as practical solutions to environmental monitoring and maritime security challenges such as illegal fishing, piracy, smuggling, and human trafficking~\cite{10341867}. To effectively address these issues, marine robots, especially unmanned surface vessels (USVs), must perform tasks such as autonomous surveillance, tracking, and intervention-based activities involving manipulation \cite{sun2021unmanned}~\cite{10406475}.


By equipping USVs with manipulators, these vessels can perform a wide range of manipulation tasks autonomously; for instance,
in maritime security, these manipulation tasks allow USVs to actively respond to incidents, such as assisting in the interception of suspicious vessels and retrieving contraband \cite{7271504,10197568,6459926,5714207}.
Similarly,
in port logistics and maritime transportation \cite{jmse11030557}, USVs with manipulators can assist in transporting cargo between docks and ships within port terminals, optimizing loading and unloading processes and reducing turnaround times for vessels, an integral part of this transportation task is the successful retrieval of objects from the dock. \cite{CORDIS_2013}

This paper addresses the challenge, where a USV equipped with a manipulator retrieves objects from the dock in the presence of disturbances due to sea waves.
The USV will experience dynamic and unpredictable movements. These disruptions will significantly increase the complexity of grasping tasks for the manipulator.
The challenges posed by wave-induced disturbances significantly impact tasks involving robotic arms on the water surface. Current research primarily focuses on stabilizing the end-effector pose of robot arms mounted on ships in the presence of ocean waves and currents, with applications such as stabilizing a UAV landing platform \cite{9636055}, not actual manipulation tasks. \cite{woolfrey2021predictive} and \cite{from2011motion} have demonstrated this using Model Predictive Control (MPC). This approach can be extended to our case; however, a significant drawback is its dependency on obtaining an accurate plant model, which is particularly challenging in maritime environments.

\begin{figure}
\begin{center}
\includegraphics[scale=0.2650]{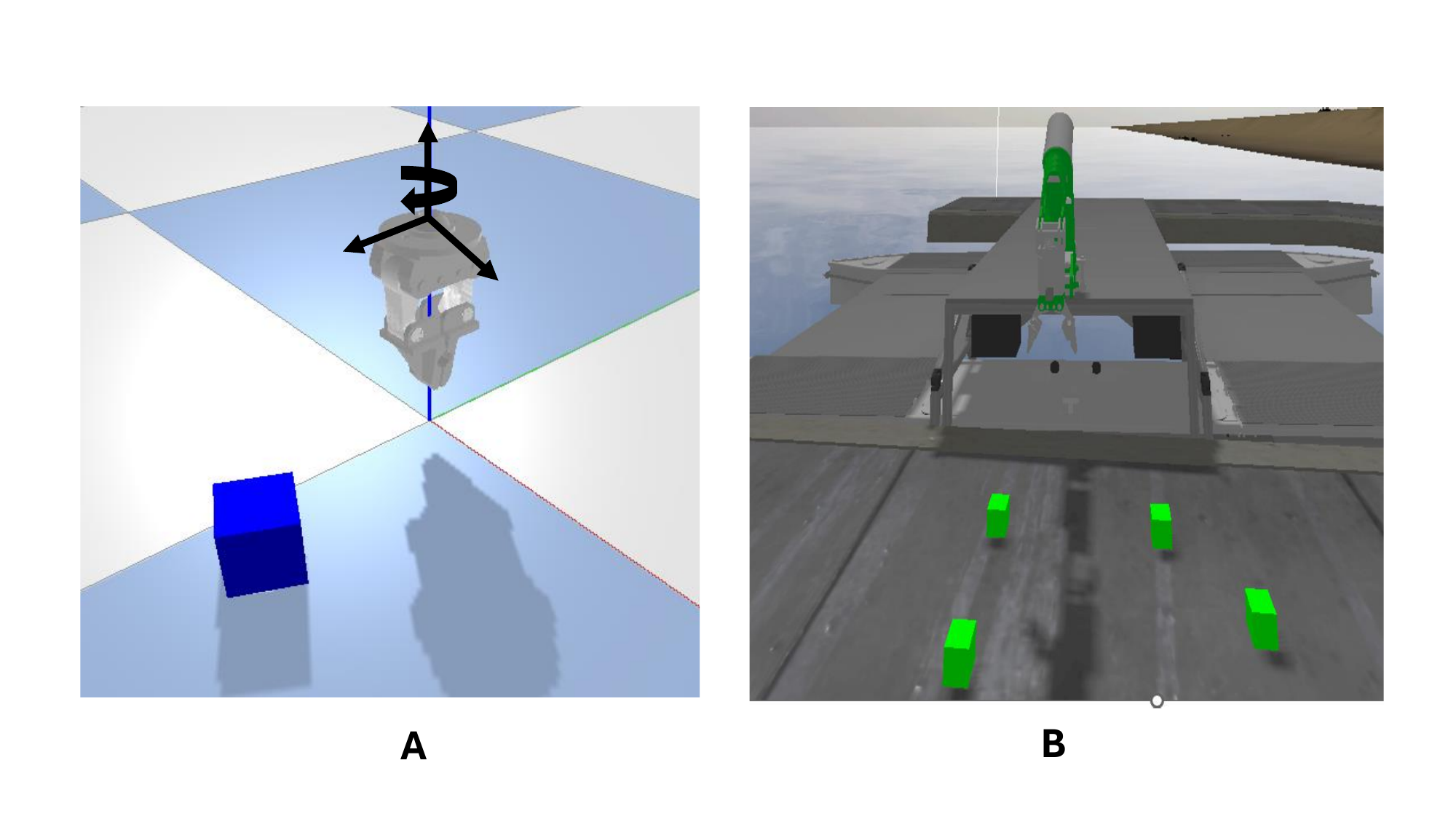}    
\caption{Training and testing environments for object manipulation. Fig-A: Pybullet training environment for object grasping. Fig-B:  shows the testing environment.} 
\label{fig:gym}
\end{center}
\end{figure}
We propose applying deep reinforcement learning (DRL) to accomplish the manipulation task, as it offers several benefits for grasping in dynamic environments. Unlike classical control techniques, DRL methods do not depend on an explicit environment model. By learning directly from interactions, DRL eliminates the need for an accurate environmental model and only requires a training environment. Additionally, DRL provides online and closed-loop control strategies, which are highly effective in resisting disturbances. These characteristics make DRL well-suited for tasks in dynamic environments, allowing for adaptive and robust grasping performance under uncertainty.

Training a DRL agent in a dynamic environment is highly sample inefficient, often necessitating millions of steps to converge \cite{amarjyoti2017deep}. However, our findings demonstrate that an agent trained in a static environment can perform satisfactorily even when facing wave-induced disturbances. This suggests that dynamic training may be bypassed in our case. The trained agent was tested in a realistic simulation environment, which included various wave scenarios mimicking real-world conditions according to the WMO Sea state code~\cite{WMO}.

In summary, the key contributions of this work are:
\begin{enumerate}
    
    \item Investigate the challenge of retrieving objects from the dock, subject to wave disturbances, using a manipulator-equipped USV, as Previous research primarily concentrated on stabilizing the end effector pose of the robot arm in the presence of waves, neglecting actual manipulation tasks.

    \item 
    
    Experiment with DRL to carry out the retrieving task. Our findings reveal that effective results can be achieved by training in a static environment, bypassing the need for explicitly training with wave disturbance signals

    \item Develop realistic test simulations for the retriving task across diverse sea conditions, following the WMO Sea State code, and conduct an in-depth comparison and analysis of the results.
    
\end{enumerate}





\section{Related Work}

Object grasping and manipulation are crucial in robotics as they enable robots to interact and modify their environment, making them valuable for a wide range of applications~\cite{S0219843619500415}. Effective object manipulation allows robots to perform complex tasks such as picking in cluttered environments, packaging, and handling delicate items with precision.  Furthermore, advancements in object manipulation are essential for developing autonomous systems that can adapt to dynamic and unstructured environments.

Manipulation in the marine environment using manipulators involves specialized robotic arms or remotely operated vehicles (ROVs) to precisely handle objects such as underwater equipment, subsea structures, and cargo.
Current research about object manipulation using manipulators in marine environments is focused on objects either floating or underwater. For instance, in~\cite{PENALVER2015201,DU2023255} vision-based grasping systems are employed that track and grasp objects.
Some studies such as~\cite{Carlucho2020ARL} used RL paradigm for controlling manipulators attached to Autonomous Underwater Vehicles in underwater environments. However, these applications are confined to scenarios with minimal relative disturbances between the manipulator and the objects.


Deep Reinforcement Learning has emerged as a promising approach for robot grasping due to its ability to handle complex and dynamic environments. It demonstrates potential in controlling robots for grasping both rigid \cite{kalashnikov2018qtopt} and deformable objects \cite{jangir2020dynamic}. For instance, \cite{buchler2022learning} demonstrated impressive results with a custom-made high-response manipulator, showcasing robotic table tennis skills learned from scratch. However, a significant challenge associated with DRL is its sample inefficiency. Training an agent often requires millions of steps in static environments \cite{RLReview}, and the demands are even higher in dynamic settings. To address this issue, \cite{fang2019dher} proposed "Dher: Hindsight experience replay for dynamic goals". This method utilizes failure episodes to generate artificial success trajectories by treating the final state as the desired goal. While effective, it currently achieves only dynamic reaching in simulations. Another strategy to improve sample efficiency involves transferring a trained agent from simulation to the real world (Sim2Real). \cite{chen2021deep} pioneered this approach, demonstrating the ability of a deep learning agent to hold a moving object in the real world. Their work trained a Baxter robot to grasp a moving object in simulation before transferring it to the real world, achieving a 7/30 success rate in straight-line scenarios. Due to the inherent complexity of dynamic grasping, direct training in dynamic environments often suffers from low success rates and extended training times. \cite{wu2022grasparl} tackled this challenge using adversarial reinforcement learning. They employed two separate agents - one for the end-effector and one for the object - where the object agent gradually increased its movement complexity. This method achieved a 3/5 success rate for grasping a mustard bottle.
An innovative approach introduced in \cite{Xu2023ImprovingRL} avoids training a dynamic RL agent directly. Instead, they train an agent to grasp a static object and then enhance its ability with a trajectory prediction module. This approach has the potential to achieve both high sample efficiency and good performance in dynamic environments. Both \cite{Xu2023ImprovingRL} and \cite{wu2022grasparl} use Soft Actor Critic (SAC) which is the current state of the art in DRL algorithms, A central feature of SAC is entropy regularization. The policy is trained to maximize a trade-off between expected return and entropy. This connection to the exploration-exploitation trade-off helps accelerate learning and prevents premature convergence to suboptimal solutions.
Recent advancements explore using meta-parameters to adapt grasping policies to different scenarios. These meta-parameters control various aspects of the grasping pipeline, such as the object pose predictor's look-ahead time and the motion planner's time budget. For instance, \cite{Jia2023DynamicGW} demonstrates a meta-controller that learns to dynamically adjust these parameters based on the current scene, leading to improved grasping success rates and reduced grasping time while also managing to grasp objects in the presence of obstacles.

This work presents a unique grasping challenge: using a reinforcement learning (RL) agent to control a robot arm that must grasp stationary objects on the dock while buffeted by ocean wave disturbances. This scenario departs from research mentioned above that focused mainly on grasping of moving objects with a static robot arm. Our approach requires the RL agent to not only account for the object's pose and gripper dynamics but also to compensate for the unpredictable movements of the robot arm caused by the waves. This presents a more complex and realistic manipulation task that pushes the boundaries of current RL grasping techniques.

\section{Methodology}

\subsection{Task Definition}

A robotic arm is mounted on an unmanned surface vessel (USV) that will encounter dynamic and unpredictable movements caused by waves and currents. These disturbances significantly increase the complexity of grasping tasks for the robotic arm. The primary objective of this study is to develop a reinforcement learning (RL) agent capable of controlling the robotic arm to autonomously retrieve objects from a docking station despite the external disturbances from the USV's movements. The RL agent must adapt to and respond to these disturbances to grasp and successfully retrieve the objects.

To simplify the task, we make the following assumptions and constraints:

\begin{itemize}
    \item Complete knowledge of the gripper and object position and orientation in the world at all times.
    \item Only top grasps are considered. 
\end{itemize}




Training the RL agent in a wave-induced dynamic environment is inefficient due to its randomness, requiring extensive training steps. Instead, we propose training the agent in a static environment to grasp randomly placed objects. Here, the agent learns to quickly close the distance between the gripper and the object to maximize rewards. If wave disturbances are slow and periodic, the agent’s rate of closing the distance should outpace the rate of these disturbances, leading to successful grasps. We hypothesize that an agent trained in a static environment will perform well in minor wave disturbances. We will train the agent statically and test it in a dynamic environment that simulates waves, demonstrating its effectiveness in our results and discussion sections.




\subsection{Reinforcement Learning Agent Training}
In RL-based approaches, an agent undergoes a learning process to make decisions by actively engaging with its environment. The agent performs actions within the environment, gains feedback through rewards or penalties, and then gradually adjusts behavior to maximize its cumulative reward.
The agent perceives the environment through observations, where each observation represents the current state of the environment; the observation space represents all possible states that the environment can assume, while the action space comprises all feasible actions that the agent can undertake. Central to RL is the reward function, a function that maps the observation-action pair to a scalar value that describes how desirable the action taken is in light of the given observation, thereby guiding the agent’s decision-making process; the policy is the function or the strategy that maps the agent’s current observation to actions, the agent goal at the end is to learn a policy that will maximize its cumulative reward.
.

The RL agent will learn to control the gripper's position and orientation to pick up a randomly placed cube on a plane and lift it to a specified height. To ensure only top grasps, the gripper has a fixed orientation with respect to the cube in two degrees of freedom (pitch and roll). 
Therefore, the agent controls the gripper's position p=\{x,y,z\} and rotation along the z-axis (yaw angle).
For the training purpose, we created a simulation environment using Pybullet as shown in Fig.~\ref{fig:gym}. The simulation environment's structure and functionality are consistent with the interface provided by OpenAI Gym, allowing for the utilization of established reinforcement learning algorithms. The RL algorithm that we selected is the Soft Actor-Critic (SAC) algorithm introduced by \cite{pmlr-v80-haarnoja18b}


\subsection{Soft Actor-Critic (SAC) algorithm}

\begin{figure}
\begin{center}
\includegraphics[scale=0.270]{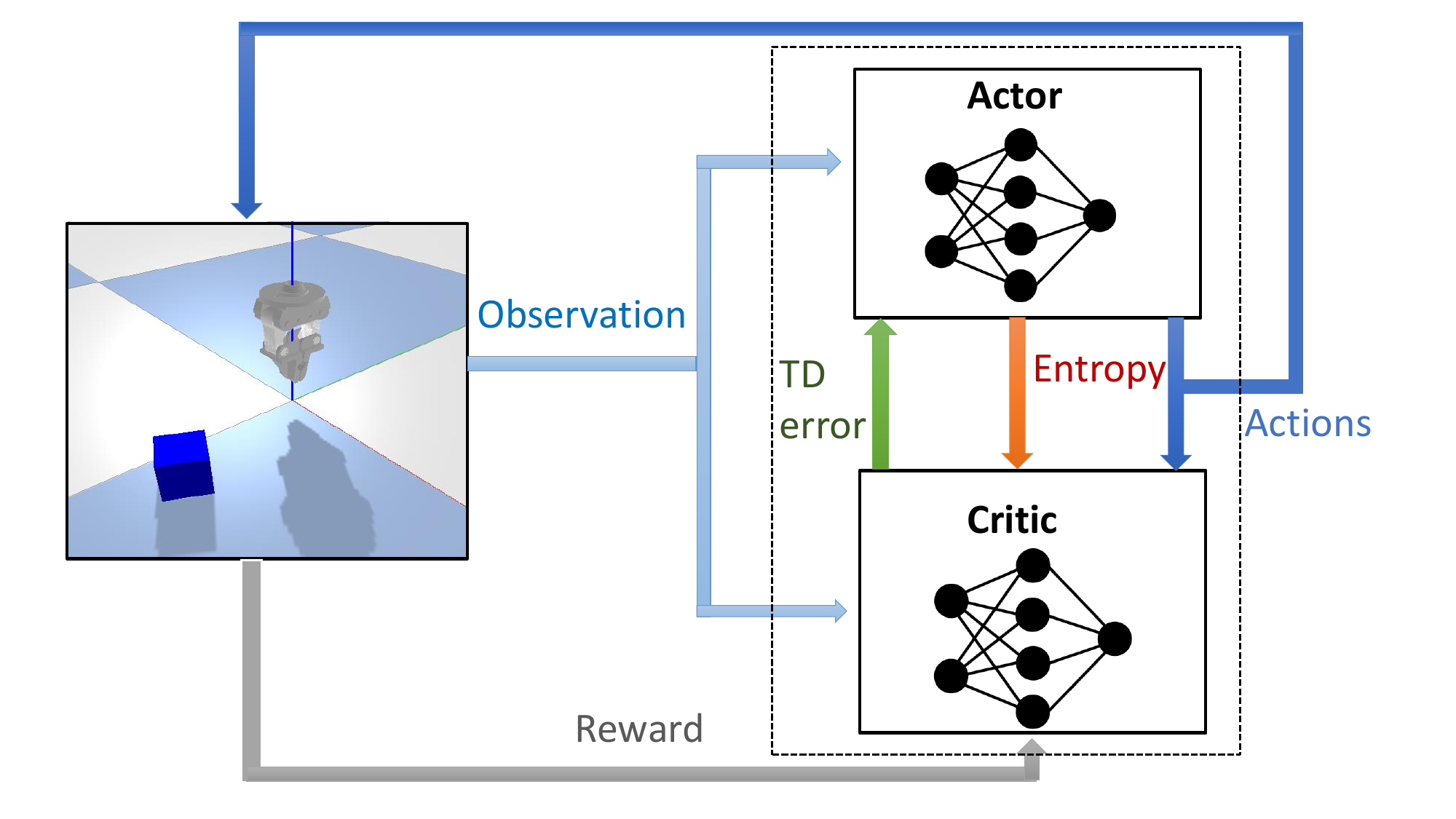}
\caption{The actor-critic framework displaying the actor and critic neural networks and the environment, the actor encapsulates the learned policy while the citric  evaluate this actor-learned policy and provide feedback to guide
the actor toward an optimal path.}
\label{fig:SAC}
\end{center}
\end{figure}

Soft Actor-Critic (SAC) is a state-of-the-art model-free  deep reinforcement learning algorithm that leverages deep neural network for solving problems with continuous action spaces. This off-policy algorithm incorporates an actor-critic framework, shown in Fig.~\ref{fig:SAC}; the actor learns a stochastic policy that determines actions based on the current state. This stochasticity allows SAC to handle uncertainties and exploration effectively. The critic evaluates this policy and provides feedback to guide the actor toward an optimal path. The goal is for the actor to learn a policy that optimizes an objective function through training.
One of the main aspects that differentiates SAC from the other actor-critic algorithms is that it tries to maximize the entropy and reward in the objective function. Therefore, the algorithm considers not only the expected rewards but also the entropy of the policy; this encourages exploration without sacrificing the pursuit of high rewards during training; equation~(\ref{eq:SAC}) describes the objective function of the SAC algorithm.

\begin{equation}\label{eq:SAC}
\begin{gathered}
\pi=\arg \max _{\pi} \underset{\tau \sim \pi}{E}\left[\sum _ { t = 0 } ^ { \infty } \gamma ^ { t } \left(R\left(s_{t}, a_{t}, s_{t+1}\right)+\right.\right. \\
\left.\left.\alpha H\left(\pi\left(\cdot \mid s_{t}\right)\right)\right)\right]
\end{gathered}
\end{equation}

$\pi$ is the policy that the actor will learn, $\gamma$ is the reward discount factor, $\tau$ is the trajectory, $R$ is the reward function, $s_{t}$ is the observation at time $t$, $s_{t+1}$ is the observations at time $t+1$, $a_{t}$ is the actions at time $t$, 
$H\left(\pi\left(\cdot \mid s_{t}\right)\right)$
 is the policy's entropy at $s_{t}$, $\alpha$ is the trade-off coefficient which controls the trade-off between the exploration and exploitation, the selection of the SAC algorithm over other RL algorithms for our case is driven by several advantageous characteristics:
 \begin{enumerate}

\item SAC is a model-free algorithm that doesn't require an explicit model of the environment dynamics. 
\item  The excellent exploration capabilities of SAC are desirable in our continuous observation and action space.
\item  SAC exhibits robustness to hyper-parameters such as learning rate and reward discount factor, making it less sensitive compared to other model-free algorithms
 \end{enumerate}
 The critic network is a Q value network, with an input layer of size 40, 3 hidden layers of size 256, and an output layer of size 1. The input to the critic is the observations and actions, and the output is the expected Q value. The Q value represents the predicted cumulative reward an agent can obtain by taking a particular action in a given state, this network is updated to minimize the error between its predictions and the observed reward. the actor-network has an input layer of size 30, which the
size of the observation, and 28 hidden layers with a size of 256, and an output layer of size 5 repressing the size of the actions, the actor aims to maximize the Q-value according to the critic, after training within the SAC framework, the actor encapsulates the learned policy. The actor then serves as the agent during testing.

The reward function $R$, action space $a$ and observation space $s$ that we used, are all defined below :

\subsubsection{Action Space $a$ }\phantom{}


 As mentioned earlier, the agent controls over the gripper's x, y, and z positions and the yaw angle. Additionally, the agent can adjust the gripper's joint angle to open and close it. The action space is continuous and is described in equation~(\ref{eq:action}).

\begin{equation}\label{eq:action}
a_{t}=\left\{\Delta x, \Delta y, \Delta z, \Delta \psi_{g}, g_{state}\right\} 
\end{equation}

all the values in the action space are normalized to [-1,1]. $\Delta x$, $\Delta y$, $\Delta z$, and $\Delta \psi_{g}$ are small changes in the gripper position and orientation with respect to the world, and the gripper control involves scaling them by a factor $\beta$ which is chosen to be 0.5 .  $g_{state}$ denotes the state of the gripper, with"-1" indicating that it is fully closed, and "1" indicating that it is fully open, $a_{t}$ represent the action taken at time $t$ by the agent.

the new gripper's position and orientation at time $t+1$ will be calculated based on the action taken at time $t$ as follows:
\begin{equation}\label{eq:x}
x_{t+1}=x_{t}+\Delta x*\beta
\end{equation}
\begin{equation}\label{eq:y}
y_{t+1}=x_{t}+\Delta y*\beta
\end{equation}
\begin{equation}\label{eq:z}
z_{t+1}=x_{t}+\Delta z*\beta
\end{equation}
\begin{equation}\label{eq:psi}
\psi_{g(t+1)}= \psi_{g(t)}+\Delta \psi_{g}*\beta
\end{equation}

\subsubsection{Observation Space } \phantom{}

 The observation space is described as follows:
 
\begin{equation}\label{eq:observation}
s_{t}=\left\{p^{cube}, p^{g},D_{cube_{-} g}, g_{state},\psi_{g}, \psi_{cube}\right\}
\end{equation}

where, the variables $p^{cube}$ and $ p^{g} $ represent the position of the cube and the gripper in the world coordinate, $g_{state}$ is the state of the gripper as in equation~(\ref{eq:action}), and $D_{cube_{-} g}$ is the distance between the gripper and the cube. 
$\psi_{g}$ and $\psi_{cube}$ denote the yaw angles of the gripper and the cube, respectively. These angles signify the rotation about the Z-axis in the world coordinate system.

$s_{t}$ represent the observation of the environment at time $t$, We stack three observations from
consecutive time steps and obtain the observation that will be used during training and testing, therefore the observation at time $t$ becomes:
\begin{equation}\label{eq:stackedobservation}
s_{t}^{stacked}=\left\{s_{t},s_{t-1},s_{t-2} \right\}
\end{equation}

this will give the agent access to information from different time steps and will allow it to capture temporal dependencies such as velocity and acceleration, and then learn the relation between the actions and such dependencies implicitly.

\subsubsection{Reward Function Design}\phantom{}



We decomposed the task into three distinct sub-tasks:  \textit{reaching}, \textit{grasping}, and \textit{lifting}.
Successful completion of the task involves the accomplishment of each sub-task. Therefore, We have designed a dedicated reward function for each sub-task.

For reaching, we have two reward functions, one related to the position and one related to the orientation. The position reward is calculated as follows :
\begin{equation}\label{eq:r1}
 1-\tanh \left(1.66 D_{cube_{-} g}\right) 
\end{equation}

the orientation reward is :
\begin{equation}\label{eq:r2}
1-\tanh a b s\left(\psi_{cube}-\psi_{g}\right)
\end{equation}
In the grasping stage, the agent receives a reward of 0.5 when both fingers of the gripper touch the cube. For the lifting stage, a reward of 2 is granted when both fingers touch the cube and the cube is lifted by 1 cm from the ground. To ensure the successful completion of lifting, the cube is to be lifted at least 20 cm off the ground, and to achieve that  an additional reward of 5 is given when the cube is grasped and at least 20 cm off ground. We employ a dense reward scheme to enable the agent to learn rapidly. At each time step, the agent's total reward is the sum of these individual rewards.

\subsection{Testing Environment}

\begin{figure}
\begin{center}
\includegraphics[scale=0.270]{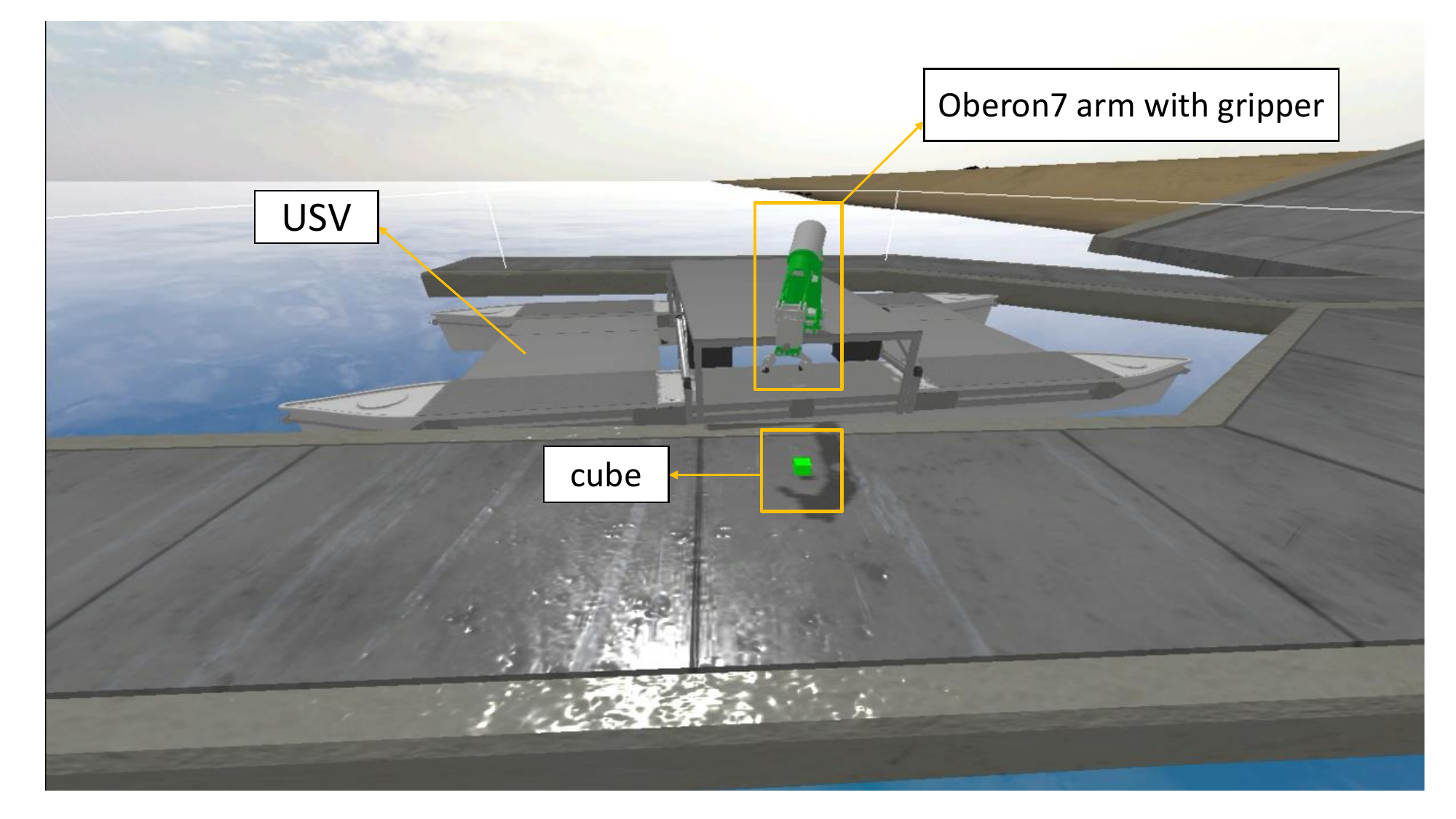}
\caption{Testing setup scene featuring an Unmanned Surface Vehicle (USV) equipped with the Oberon7 arm and its gripper, and the object to be picked.}
\label{fig:MBZIRC}
\end{center}
\end{figure}

To evaluate the performance of the developed agent in achieving its primary objective of retrieving the objects from the docking station by controlling the robotic arm mounted on the Unmanned Surface Vehicle (USV),  we employed the MBZIRC simulator developed by \cite{MBZIRC} . It is an open-source marine simulator developed for the Mohamed Bin Zayed International Robotics Challenge (MBZIRC). It is developed using C++, Python, and ROS2 Galactic which is developed by \cite{ros2} and used for communication among the components. The simulator uses Gazebo Ignition developed by \cite{ignition} for hydro-static and hydrodynamic simulations. The testing setup is illustrated in Fig.~\ref{fig:MBZIRC}, which includes the USV equipped with the Oberon7 arm, featuring the same gripper used during training. Additionally, the cube on the dock is the same size as the one used in the training phase.
The MBZIRC simulator allows the simulation of various sea states. A sea state characterizes the condition of the water surface, encompassing wave height, period, and current speed. These parameters are based on statistical data collected from real-world seas. This feature enables us to assess the agent's performance in an environment resembling real-world scenarios, providing a robust evaluation of its capabilities.

To enable the agent to control the robotic arm within the simulator, we recreate the observation space on which the agent was trained using information obtained directly from the simulation. We didn't implement any sensors in the simulation to get this information, as we can access the ground truth information directly using ROS topics, we then provide these observations to the agent, allowing it to generate the necessary actions. The changes in position and orientation obtained from the actions are scaled by the factor previously used during agent training. These scaled actions represent the change in the gripper pose.  It is added to the current gripper pose, yielding the next gripper pose with respect to the world as shown in equations ~\ref{eq:x},~\ref{eq:y},~\ref{eq:z} and ~\ref{eq:psi}, since we're considering top grasps we need to ensure that the gripper is always pointing towards the ground, therefore the roll and pitch angles with respect to the world are always chosen to be $\phi=0$  and $\theta=0.5*\pi$  respectively.

The next obtained pose is then input into an inverse kinematics solver. This solver works on an optimization problem to determine the joint angles that best achieve the specified pose. This calculation is performed based on the current pose of the robot’s base.

\begin{figure}
\begin{center}
\includegraphics[scale=0.270]{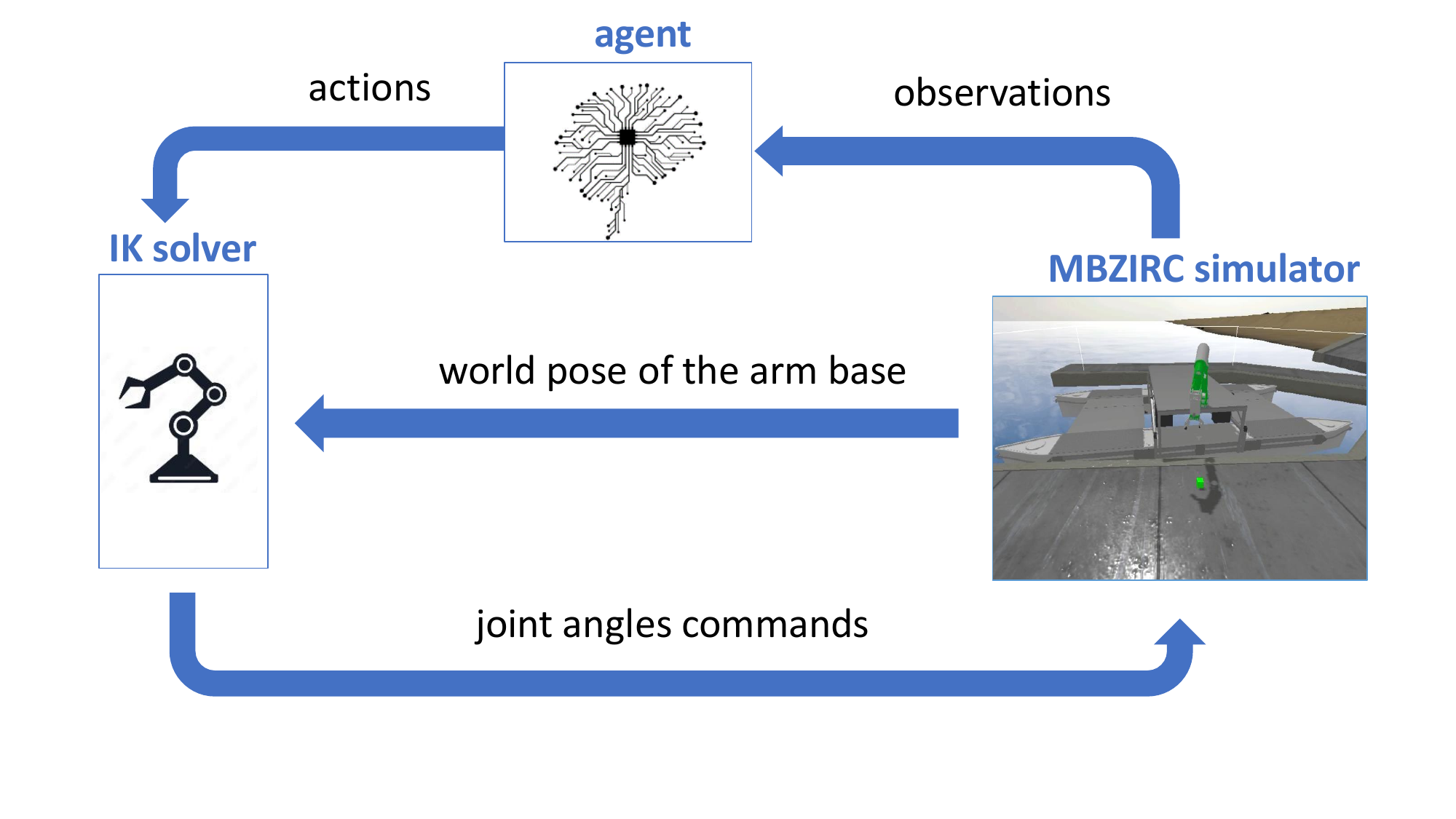}
\caption{ The figure depicts the dynamic interaction between the agent, the simulator, and the inverse kinematics solver at each time step, highlighting the process and sequence of command exchanges.}
\label{fig:flow}
\end{center}
\end{figure}

Additionally, based on the gripper state obtained from the action, we determine the required gripper joint angle, The resulting joint angles of the arm and the gripper are then sent to the simulator via ROS2 messages as joint angle commands, A PID controller executes these commands within the simulation, allowing the arm to respond and move accordingly, Fig.~\ref{fig:flow} depicts the overview of this process.

\section{Results and Discussion}
\subsection{Training Results}
As outlined in Section 2, we employed the soft actor-critic (SAC) reinforcement learning algorithm for training. In our implementation, we used a variation that enables the automatic tuning of the alpha parameters throughout the training process. This adaptive adjustment optimizes the trade-off between exploration and exploitation. Table ~\ref{tb:parameters} provides a detailed overview of the hyper-parameters utilized during the training phase.The hyper-parameters used were similar to the ones used in \cite{zhan2021framework}.

\begin{table}[hb]
\begin{center}
\caption{Hyper-parameters of the SAC algorithm.}\label{tb:parameters}
\begin{tabular}{|c|c|c|c}
\hline
hyper-parameter & value \\ \hline
$\gamma$  & 0.98 \\
learning rate & 0.0001  \\ 
 initial $\alpha$ & 0.5 \\ \hline
\end{tabular}
\end{center}
\end{table}

\begin{figure}
\begin{center}
\includegraphics[scale=0.270]
{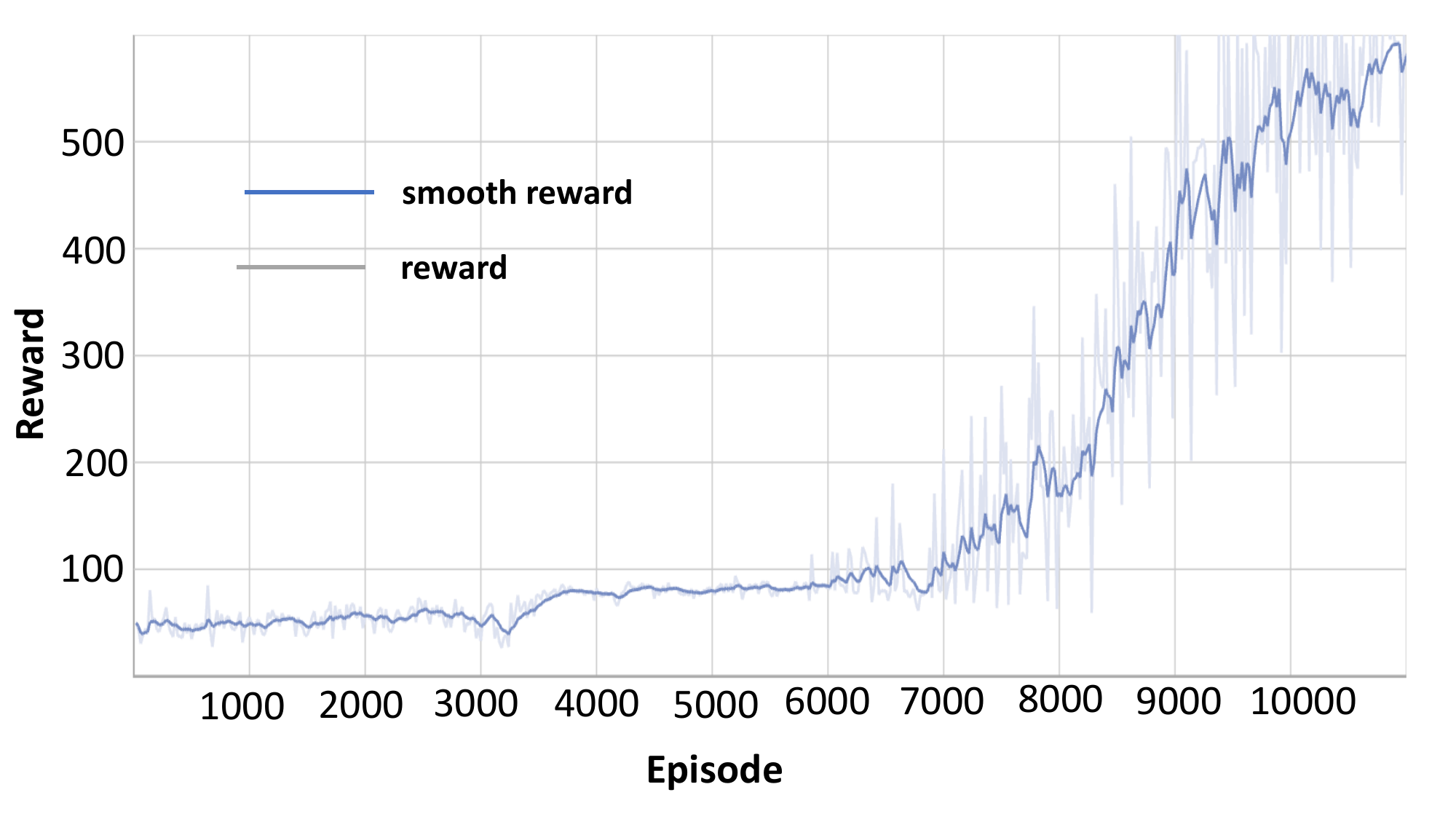}
\caption{The reward improvement during training, The original reward is represented by the grey line, whereas the smoothed reward, achieved through a window of size 5, is illustrated by the blue line.}
\label{fig:training}
\end{center}
\end{figure}

Each training episode consists of 100 time steps, and we employed a dense reward scheme, meaning that the reward is computed at each time step. As shown in Fig.~\ref{fig:training}, the training process required approximately 10,000 episodes to converge, yielding an agent with good performance, as evidenced by the smooth reward, which is close to the maximum reward. The task is considered successful if, by the conclusion of episode, the agent has successfully grasped and lifted the cube at least  20 cm off the ground. Our trained agent achieved this criterion 93 percent of the time, demonstrating satisfactory performance in the training environment.
\subsection{Testing Results}

By employing the MBZIRC simulator, we established the testing environment detailed in Section 3 and illustrated in Fig.~\ref{fig:MBZIRC}. This environment is utilized to assess our trained agent's capability to effectively grasp and retrieve objects from the dock station under wave disturbances. We conducted tests under three different sea states classified as 0, 1, and 2, following the codes designated by The World Meteorological Organization, which can be found in  \cite{WMO}. A comprehensive description of these sea states is provided in table~\ref{tb:seastate}.
During testing in sea state one, we employed a wave amplitude of 0.1 meters, representing the maximum amplitude for this sea state. The corresponding wave period was set to 5 seconds. In sea state two, we utilized a wave amplitude of 0.5 meters, the maximum for this sea state, and maintained a consistent wave period of 5 seconds. This uniformity ensures consistency and facilitates the comparisons of the agent performance in different sea states.

\begin{table}[hb]
\begin{center}
\caption{WMO Sea State Code.}\label{tb:seastate}
\begin{tabular}{|c|c|c|}
\hline
WMO Sea State Code & Wave amplitude(m)& Characteristics  \\ \hline
 0 & 0 & Calm(glassy) \\
  1  & 0 to 0.1 &Calm(rippled) \\ 
  2 & 0.1 to 0.5&Smooth (wavelets) \\ \hline
\end{tabular}
\end{center}
\end{table}

Similar to the training environment, successful completion of the task in the testing environment is defined by the agent's ability to grasp the cube from the dock and lift it a minimum of 20 cm within 30 seconds of simulation time. Unlike during training, we extend the time window in this scenario, acknowledging that the agent faces additional challenges in overcoming wave disturbances.
For each sea state, we did 15 tests and calculated the success rate, which is shown in table~\ref{tb:succesrate}.

\begin{table}[hb]
\begin{center}
\caption{Agent success rate under different sea states.}\label{tb:succesrate}
\begin{tabular}{|c|c|c|c}
\hline
Sea State Code & Agent success rate \\ \hline
 0 & 0.933\\
  1  & 0.87 \\ 
  2 & 0.8\\ \hline
\end{tabular}
\end{center}
\end{table}
In sea state 0, characterized by the absence of wave disturbances, the agent achieves a success rate of 0.933, which is very similar to the success rate observed in the training environment, which was 0.93. This alignment is expected since no disturbances are applied to the gripper in the training and sea state 0 testing environments. in general the results indicate that the agent performs well even when faced with wave disturbances despite not being trained specifically for that purpose. and this behavior can be attributed to the following :

\begin{enumerate}
   \item The wave disturbances primarily affect the gripper,
not the object which remains stationary on the dock.
Given that, the object is considered static with respect to the world, providing observations akin to
the agent’s training, and since RL control operates in
an online and closed-loop manner, the agent continuously receives feedback from the environment and
can immediately adjust its actions. Consequently, the
agent persistently drives the gripper to approach the
object despite disturbances attempting to push it
away.

    \item The agent can move the gripper quickly to reach the object. Reaching is a necessary part of achieving the grasping task; therefore, the agent learned during training how to control the gripper to efficiently close the distance to the object as fast as possible to maximize its reward; the agent's ability to move quickly resulted from it implicitly learning representations of velocity and acceleration, and associating them with specific actions to optimize the reward, The agent's capacity to comprehend velocity and acceleration without them being explicitly shown in the observations is attributed to our practice of stacking observations from multiple time steps and use it in training.
    \item  The slow and periodic nature of the sea wave disturbances causes relatively slow fluctuations in the distance between the object and the gripper. When the waves decrease this distance, it is advantageous for the agent since it brings the gripper closer to the object. Therefore, the waves won't always work against the agent.
\end{enumerate}
    
\begin{figure}
\begin{center}
\includegraphics[scale=0.270]
{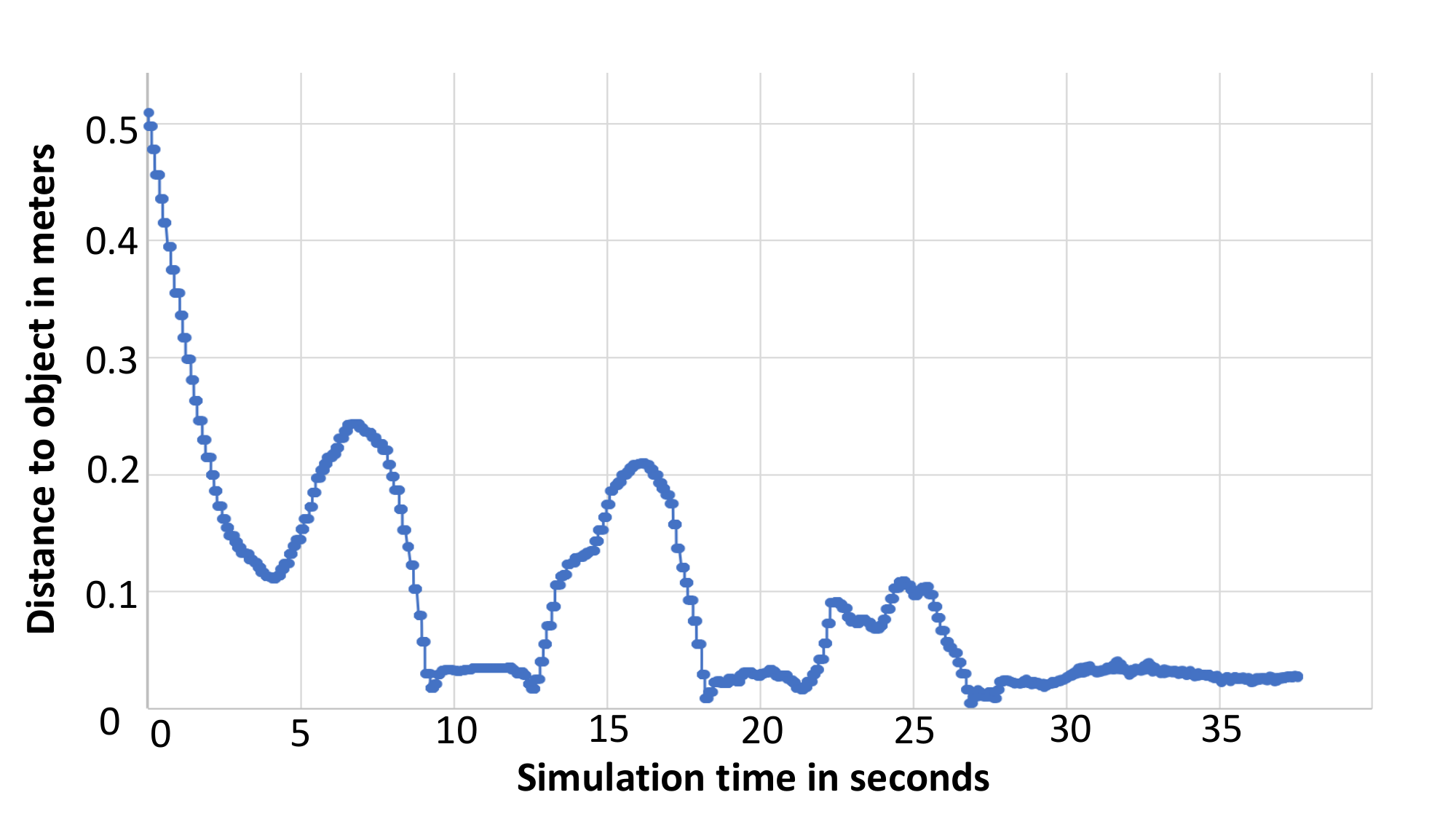}
\caption{The change in the distance from the gripper to the object versus time. It shows how the agent approaches the object despite the wave-induced disturbances.}
\label{fig:distance}
\end{center}
\end{figure}

\begin{figure}
\begin{center}
\includegraphics[scale=0.270]
{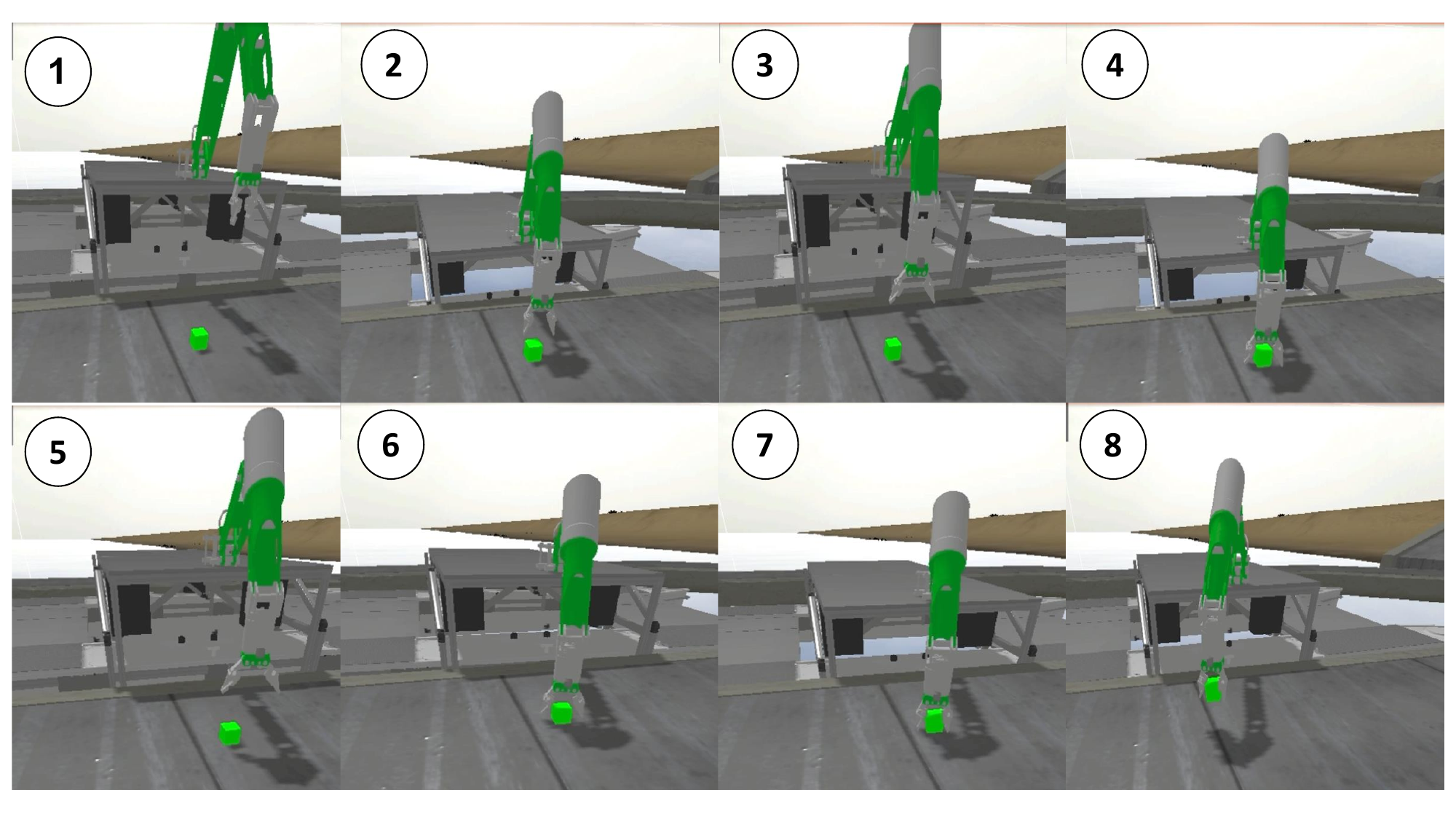}
\caption{ Sequence of snapshots of the execution of the object retrieval task. In screenshot four, the arm almost grasped the object, but in screenshot five, the arm gets pushed away due to the waves; however, eventually, it manages to resist the waves and grasp the object.}
\label{fig:screenshots2}
\end{center}
\end{figure}

As a result of these factors mentioned above, the agent effectively outpaces the slow and periodic disturbances of the waves, gradually minimizing the distance to the object over time, ultimately leading to a successful grasp. This is evident in Fig.~\ref{fig:distance}
, where the peaks in the distance to the object gradually diminish over time. This reduction signifies that the agent is minimizing the distance to the object. Eventually, we observe the distance stabilizes, indicating the agent's successful grasp of the object. A visual representation of the arm movement to grasp the object is evident in Fig.~\ref{fig:screenshots2}, where we can see screenshots of the arm movement to grasp the object; we can see in screenshot four that the arm is almost grasping the object, but in screenshot five, it gets pushed away by sea waves. However, it manages to resist the disturbances and eventually successfully grasps the object.

In sea state one, the success rate is similar to zero state due to low wave amplitude and long periods, resulting in small and slow disturbances. In sea state two, with increased amplitude but the same period, the agent's success rate drops significantly due to the increased speed of disturbances, making it more difficult for the agent to resist and minimize the distance within the allocated time to grasp the object successfully.

\section{Conclusions}

In this work, we addressed the challenging issue of retrieving objects from a dock using a USV equipped with a robotic arm.  we trained a reinforcement learning agent based on the actor-critic framework and performed testing in a realistic simulation using wave characteristics based on the WMO Sea state code, which showed that when training an RL agent to grasp randomly placed objects in static environment it acquires the ability to resist disturbance and can effectively handle wave-induced disturbances without specific dynamic training.

Our proposed method solely focuses on top grasps and accounts for wave-induced disturbances only, overlooking drifting. Drifting refers to the USV moving due to currents; unlike waves, drifting presents non-periodic disturbances which pose a challenge for the agent to address. Moving forward, we aim to enhance the agent's capabilities to incorporate more complex scenarios such as mobile manipulation for marine intervention. Furthermore, we will test the developed approach in a real sea environment with an underwater manipulator.

\begin{ack}
This work is supported by the Khalifa University of Science
and Technology under Award No. MBZIRC-8434000194,
CIRA-2021-085, FSU-2021-019, RC1-2018-KUCARS. 
\end{ack}

\bibliography{ifacconf}

\end{document}